\def\year{2018}\relax
\documentclass[letterpaper]{article} 
\usepackage{aaai18}  
\usepackage{times}  
\usepackage{helvet}  
\usepackage{courier}  
\usepackage{url}  
\usepackage{graphicx}  
\begin{document}
%
\title{From Algorithmic Black Boxes to Adaptive White Boxes:\\ Declarative Decision-Theoretic Ethical Programs as Codes of Ethics}
\author{Martijn van Otterlo\\
Vrije Universiteit Amsterdam, The Netherlands\\
}
\maketitle
\begin{abstract}
Ethics of algorithms is an emerging topic in various disciplines such as social science, law, and philosophy, but also artificial intelligence (AI). The value alignment problem expresses the challenge of (machine) learning values that are, in some way, aligned with human requirements or values. In this paper I argue for looking at how humans have formalized and communicated values, in professional codes of ethics, and for exploring \emph{declarative decision-theoretic ethical programs} (\textsc{DDTEP}) to formalize codes of ethics. This renders machine ethical reasoning and decision-making, as well as learning, more transparent and hopefully more accountable. The paper includes proof-of-concept examples of known toy dilemmas and gatekeeping domains such as archives and libraries.
\end{abstract}


\section{Introduction}
Imagine you get a message on Facebook saying \emph{"Hi there... we have computed an above-average depression score for your friend. Based on other data, we know that positive actions by close friends can have an impact on his score. If you want to help your friend, please consider using more smileys and posting more frequently. Thank you!"} This may sound \emph{creepy} \cite{tene:creepy}, but it fits in recent efforts by Facebook to predict potential suicides~\footnote{https://www.wired.com/2017/03/artificial-intelligence-learning-predict-prevent-suicide/}, and by Google to detect depression~\footnote{https://www.theguardian.com/commentisfree/2017/aug/25/google-clinical-depression-privacy}. Besides invoking people's social networks, one could also limit their access to potentially harmful information, in essence implementing forms of \emph{censorship}~\footnote{http://fortune.com/2017/05/22/facebook-censorship-transparency/}. Such decisions too are becoming more common, for example by Facebook to fight terrorism~\footnote{http://www.telegraph.co.uk/news/2017/06/16/facebook-using-artificial-intelligence-combat-terrorist-propaganda/}, by Google to battle fake news~\footnote{http://www.telegraph.co.uk/news/2017/06/16/facebook-using-artificial-intelligence-combat-terrorist-propaganda/} and by Twitter's new tools~\footnote{https://www.forbes.com/sites/kalevleetaru/2017/02/17/how-twitters-new-censorship-tools-are-the-pandoras-box-moving-us-towards-the-end-of-free-speech}. With filter bubbles, fake news and social bots, platforms such as Facebook and Google may need to do something, but they often result in situations which humans immediately value differently, such as in the removal~\footnote{https://www.theguardian.com/technology/2016/sep/09/facebook-reinstates-napalm-girl-photo} of the iconic "napalm girl" photo due to Facebook's anti-nudity policy.

A novel development in such \emph{algorithmic decision making} is the tension between the dependence of humans on such services for information consumption, and the often intransparent, \emph{black box} nature of algorithmic decisions. The \emph{algorithmization} of society brings us many novel ethical issues in cases ranging from suicide prevention to autonomous cars \cite{goodall:crashes}. The new field \emph{ethics of algorithms} \cite{mittelstadt:survey,otterlo:documentalist} goes beyond classical privacy and surveillance concerns and broadly studies the impact of algorithms, including \emph{fairness}, \emph{accountability} and \emph{transparency}. Recent constructive advances in AI focus on incorporating (ethical) values into systems through \emph{value alignment}: how to \emph{"ensure that their behavior is aligned with the interest of the operators"} \cite{taylor:alignment}. In this paper I propose a novel way to address value alignment inspired by "pre-algorithmic" \emph{code of ethics}, in which humans encode ethical norms and values of a \emph{profession}. By formalizing existing human norms into \emph{declarative decision-theoretic ethical programs} (\textsc{DDTEP}) one can i) reason and learn using general, high-level decision models, ii) employ and extract domain knowledge, and iii) solve the value alignment problem by starting with human-agreed ethical values in a domain, and only learn additional domain-specific knowledge. This paper explores this novel approach and highlights new research directions.

The outline of this paper is as follows. In the coming sections I describe human formalizations of ethics and the ethics of algorithms, and I argue how expressive logical models of ethics address concerns raised by algorithms and typical properties of professional codes of ethics. Afterwards I solve elements of previously introduced toy ethical domains and illustrate how decision-theoretic logic could be used to implement human ethical values in a transparent way in the archival domain. This paper ends with a list of open research directions, technical and domain-specific.


\section{Ethics and Human Values}

A different take on the previous Facebook suicide detection case is the work by \citeauthor{juznic:suicide}\ \shortcite{juznic:suicide} who performed a "mystery shopper" experiment in public libraries. They approached librarians with morally laden topics such as necrophilia, photos of dead people, and information about \emph{"how to commit suicide"}. Interestingly, most librarians were not shocked at all by the requests and treated them professionally as pure information enquirements. \emph{"Our conclusion was that the librarians in public libraries satisfied the need for information as much as they felt inclined to do so, and this was not affected by judgments about the ethical status of the required item of information”} \cite{juznic:suicide}. A key difference is that Facebook algorithmically predicts "an interest" for suicide with the intent of preventing actual suicides, whereas the librarians merely interpret this interest as a request for information. Another difference is how the ethical values and decision procedures come about: whereas Facebook decides unilaterally and only partially discloses its intentions (which are always also connected to its profit making business model), the librarians have simple, open ethical guidelines on how to act (and may for some even cause ethical issues because they take them so strictly).

Taking (practical) action based on moral values is the domain of \emph{ethics} \cite{laudon:ethics,kizza:ethics}. \citeauthor{kizza:ethics}\ (\citeyear{kizza:ethics}): \emph{"Morality is a set of rules for right conduct, a system used to modify and regulate our behavior."} Close ties with \emph{law} exist since when a society finds certain moral values important, it can formalize such values in a law and regulate appropriate behaviors. As \citeauthor{laudon:ethics}\ (\citeyear{laudon:ethics})\ defines it: \emph{"ethics is about the decision making and actions of free human beings. When faced with alternative courses of action or alternative goals to pursue, ethics helps us to make the correct decision}. If there are options what to do, then ethics is concerned with practical reasoning about “good” and “bad” actions. Important subsequent questions are then, \emph{for whom} is something good or bad, and \emph{by who's standards}? Different answers to those questions induce a variety of ethical reasoning frameworks, with two main dimensions. One is about \emph{rules} vs. \emph{consequences}: to find the right decision one may obey rules like \emph{"thou shalt not kill"}, or look at the actual consequences and decide, e.g. to ignore the maximum speed at night when there is less traffic. The second dimension deals with \emph{"for whom"} something is good: the \emph{individual}, or the \emph{collective}. In this paper I focus on \emph{utilitarian} ethics, which is a \emph{collective consequentialist} framework aimed at maximizing the average~\footnote{Which could be bad for specific individuals though.} \emph{"goodness"} for all those affected.

Humans use several ways to enstate and enforce ethical norms. As said, the law is one option to ensure compliance, but in the digital age legal advances can be too slow to keep up with technology \cite{tene:creepy}, although recent progress has been made in the \emph{general data protection regulation} act (GDPR)~\footnote{http://www.eugdpr.org/}. A more typical way to deal with ethical norms is to formalize them as public \emph{guidelines} or \emph{rules}, with well-known examples: Asimov's three rules for robotics, the Bible's ten commandments and the \emph{Universal Declaration on Human Rights} \cite{otterlo:amnesty}.

\begin{figure*}[t]
    \centering
    \includegraphics[width=0.55\linewidth]{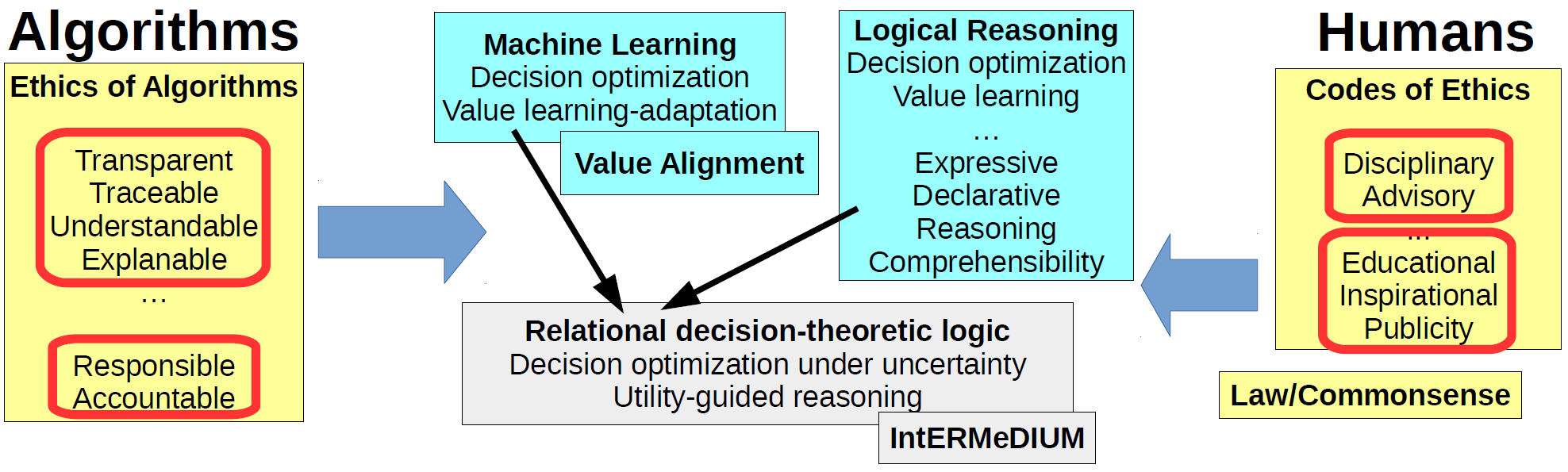}
    \caption{\textsf{Reconciling human and machine ethics through decision-theoretic logic.}}
    \label{fig:lists}
\end{figure*}

Lately the \emph{self-driving car} is the archetypical example for practical machine ethics \cite{goodall:crashes}, exemplified in Thomson's \emph{trolley problem}~\footnote{https://en.wikipedia.org/wiki/Trolley\_problem} which contains a choice between either killing five people strapped to a rail, or saving these five and killing one by pulling a lever diverting the trolley to a track with a single person (who is then killed). Trolley problems illustrate the life-or-death decisions autonomous cars may have to make. Recent empirical tests of such dilemmas suggest that humans employ one-dimensional life scales, where all outcomes (deaths) can be compared in the same scale, although time pressure affects consistency \cite{sutfeld:road}. A related study reveals that people \emph{"approved of utilitarian autonomous vehicles (that is, that sacrifice their passengers for the greater good) and would like others to buy them, but they would themselves prefer to ride in ones that protect their passengers at all costs".} Regulations here would possibly hinder the widespread acceptance of (utilitarian) self-driving cars. Thus, this domain involves clear-cut life-and-death decisions amenable to utilitarian modeling, but induces challenges when distributing the decisions' negative effects.

A more complex, and underexplored, domain is that of \emph{gatekeeping} professions \cite{otterlo:libraryness} such as 
archives and libraries \cite{otterlo:algivist,otterlo:documentalist} which have much in common with modern platforms such as Google and Facebook in terms of archival, selection and provision of information. Decisions in archives deal with who gets access to which (kind of) information, and induce ethical dilemmas involving stakeholders such as users, archivist, and persons occurring in archived materials. Typical dilemmas involve privacy, freedom of information access, and intellectual property. For example, in one of the $86$ (empirical) cases \citeauthor{ferguson:ethics}\ (\citeyear{ferguson:ethics}) list, digitization of a mass observation project archive from the 1960s/70s causes a privacy problem for individuals involved. \citeauthor{danielson:access}\ (\citeyear{danielson:access}) introduces the dilemma of \emph{equal intellectual access}, when ease of access to information is different for individuals. For example, if archival search time is costly, researchers with more resources have an advantage. In addition, archivists may choose to provide more or less assistance, for example based on a judgement of the researcher's quality, thereby making access unequal. The typical way to resolve ethical issues in gatekeeping domains is a \emph{code of ethics}, which specifies rules and values for members of the profession \cite{kizza:ethics,otterlo:documentalist}. Examples in the \emph{Society of American Archivists} $2012$ code~\footnote{https://www2.archivists.org/statements/saa-core-values-statement-and-code-of-ethics} are: \emph{"Archives are made accessible to everyone, while respecting the pertinent laws and the rights of individuals, creators, owners and users"} and \emph{"Archivists endeavour to inform users of parallel research by others using the same materials"}. Such rules are less clear-cut, and more open to interpretation but do give direction to how ethical dilemmas should be resolved. Many professions have several codes of ethics.


\section{The Ethics of Algorithms}

Turning to algorithms, ethical analysis has only started fairly recently, see \citeauthor{mittelstadt:survey}\ \citeyear{mittelstadt:survey} and \citeauthor{otterlo:documentalist}\ \citeyear{otterlo:documentalist} for pointers. In contrast to popular belief, algorithms are not \emph{objective} simply because they are mathematical. Instead, algorithms are heavily \emph{biased} by \emph{political views}, \emph{design processes} and many other factors \cite{bozdag:bias,otterlo:profiling}. Characterizing the \emph{ethics of algorithms} is hard since algorithms and potential consequences are so diverse, and situations may change over time. \citeauthor{mittelstadt:survey}\ (\citeyear{mittelstadt:survey}) define concerns about how algorithms transform \emph{data} into \emph{decisions}. Evidence can be inconclusive, inscrutable or misguided and this can cause many ethical consequences of actions, relating to \emph{fairness}, \emph{opacity}, \emph{unjustified actions}, and \emph{discrimination}. Overall, algorithms have impact on \emph{privacy} and can have \emph{transformative effects} on \emph{autonomy}, i.e. the ability for humans to make their own choices.  

Another way to structure the space of algorithms and and ethical impact, is by looking at \emph{agency}, i.e. \emph{what they are capable of}, which results in a taxonomy~\footnote{Developed in my course \emph{"ethics of algorithms"}, http://martijnvanotterlo.nl/teaching.html} with five broad classes of algorithms \cite{otterlo:documentalist}. The first type consists of algorithms that \emph{reason, infer} and \emph{search}. They employ data \emph{as it is}. The more complex they are, the more information they can extract from that data. Examples include translation, language understanding, and image recognition. Ethical concerns about such algorithms are typically about \emph{privacy} since more ways become available to interpret and link more kinds of data. A second class \emph{learns} and finds \emph{generalized patterns} in data. They are typically \emph{adaptive} versions of the first type, e.g. a scene recognition algorithm that is trained on an image stream. They introduce ethical challenges simply because they learn (outcomes are not stable), because they can statistically predict \emph{new} information (privacy), and they may severely impact users' autonomy by profiling and personalization. The third type are algorithms that \emph{optimize} to find the "best" actions. These typically employ reward functions that represent what are good outcomes and generally \emph{rank} things ("the best pizza around") or people (e.g. on Tinder). By repeatedly employing actions and optimization steps, algorithms can \emph{experiment} to find a best \emph{policy} in stochastic or unknown environments \cite{rlsota}. The ethics of experimentation has many aspects \cite{otterlo:walden}, but important is the choice of reward function (who decides has great power). The fourth and fifth classes concern \emph{physical manifestations} (e.g. robots) and \emph{superintelligence} and are out of scope here.

These five groups illustrate the many sides of the ethics of algorithms. Each comes with its own set of \emph{capabilities} but also \emph{biases}, which determine how it makes choices and which ethical challenges arise. Opening up algorithmic black boxes by making these biases, and underlying business models, transparent, can provide a way to construct AI systems that have values \emph{aligned} with human ones, and which are trustworthy, responsible, and accountable.


\section{Algorithms with Human Ethics}

The previous two sections have highlighted aspects of \emph{human} and \emph{algorithmic} ethics. On the left in Figure~\ref{fig:lists} is a partial list of \emph{requirements} from the literature on algorithmic systems that mainly have to do with \emph{"turning on the light"} \cite{tene:creepy}, i.e. transparency. These can help in making AI systems more \emph{responsible} or \emph{accountable} \cite{diakopoulos:accountability}. On the right we find the reasons why humans construct \emph{codes of ethics} for a particular domain \cite{otterlo:documentalist}. \emph{Disciplinary} and \emph{advisory} motives cover the aspects to evaluate humans in the profession, whereas the other three, \emph{publicity} in particular, deal with \emph{communication} to outside the profession to make clear which ethical behavior can be expected from members of a profession. All motives make ethical reasoning in the profession transparent, by explicitly spelling out norms and desired behaviors. Now, in the ethics of algorithms \emph{biases}, and especially a lack of transparency concerning their presence or influence on decisions, seem to be the prime source of ethical challenges. However, for human codes of ethics this is quite the opposite: a code actually \emph{is supposed to be} a fully transparent bias on the behavior of professionals. Codes of ethics can be \emph{prescriptive} (prescribe do's and don'ts) or \emph{aspirational} (only specify ideal results), which makes them more flexible than, but fairly similar to, \emph{legal} frameworks.

One intuitive way to obtain algorithmic systems that obey human values and norms is by \emph{learning} \cite{abel:rlethics} as a way to obtain \emph{value alignment} \cite{taylor:alignment}. Algorithmic systems can try to learn the values in a domain from humans, for example using \emph{inverse reinforcement learning}. For domains such as autonomous cars these values may be uncovered (e.g. see \citeauthor{sutfeld:road}\ (\citeyear{sutfeld:road}) but it is challenging for more complex domains. An advantage of machine learning is that it can cope with the uncertainty in ethical domains and that ethical behavior and values can be induced from suitable behavioral data. However, a downside of typical methods is that for challenging domains their \emph{knowledge representation} capabilities are too limited to capture a rich variety of structured knowledge domains such as ours. For that we need to turn to more expressive formalisms such as first-order and relational logic, which have been used before for \emph{machine ethics} \cite{anderson:machineethics} but often such systems are limited to symbolic reasoning and lack mechanisms to explicitly compute with uncertainty and utility. To get both, I propose to look at expressive formalisms for ethical reasoning in which \emph{knowledge} about a problem can be injected, or extracted after learning, and which can handle \emph{decision-theoretic} concepts. Various learning techniques exist for expressive formalisms, including combinations of Bayesian networks and relational logic \cite{deraedt:logical}. They can make learning more \emph{comprehensible} \cite{srinivasan:ilp} and increase \emph{explanatory power} of induced theories, simply because declarative knowledge employed can be looked at and analyzed. Some formalisms incorporate \emph{reward-based} methods (e.g. \emph{reinforcement learning}) and support decision-theoretic decision-making in expressive formalisms \cite{otterlo:relationalsurvey}.

Now, a potentially powerful AI combination of mentioned aspects can be introduced here (based on the \textsc{IntERMeDIUM} research strategy introduced by \citeauthor{otterlo:documentalist}\ (\citeyear{otterlo:documentalist}). In order to obtain value alignment in our gatekeeping domain, instead of trying to learn from scratch, we could make use of the existing bias which is provided by the codes of ethics and formalize them in suitable computational logics. This would require a combination of (expressive) declarative knowledge, value optimization, utilitarian-style ethics and learning, and could induce a baseline system that behaves roughly according to human norms. Afterwards, additional finetuning of values, domain knowledge and probabilistic aspects can be done from data and interaction. I will now illustrate this novel idea by solving in decision-theoretic logic several toy examples in ethical reasoning and a small gatekeeping problem. My approach fits in recent discussions on using AI itself to assist in ethical reasoning \cite{etzioni:ethicsbots} and related \emph{rational} approaches to machine ethics \cite{goodall:crashes}.


\section{Decision-Theoretic Logical Ethics}
\newcommand{\ttt}[1]{\begin{scriptsize}$\mathtt{#1}$\end{scriptsize}}
\newcommand{\vv}{\vspace{-0.4em}}
Let us look at some examples of \emph{declarative, decision-theoretical ethical programs} (\textsc{DDTEP}). I employ \textsc{DT-Problog} \cite{broeck:dtproblog}, which is a relational, probabilistic programming language, extended with \emph{decision-theoretic} constructs to compute with values under uncertainty. \textbf{Given} a set of \emph{action choices} (denoted \begin{scriptsize}\texttt{?::choice-1;...;choice-n}\end{scriptsize}), \emph{probabilistic dependencies} $P(a| b_1,\ldots b_n)$ (denoted \begin{scriptsize}\texttt{P::a :- b1, b2, ..., bn}\end{scriptsize}), \emph{background knowledge definitions} if $b$ and $c$ then $a$ (denoted \begin{scriptsize}\texttt{a :- b, c}\end{scriptsize}), and reward specifications $R(e) = r$ (denoted \begin{scriptsize}\texttt{utility(e,r)}\end{scriptsize}), \textbf{compute} the best action, i.e. for which the total expected reward is maximized. Solutions are computed by generating all \emph{possible worlds} modeled by the program, compiling them into an efficient data structure (e.g. \emph{algebraic decision diagrams} and others), and computing distributions and values on this datastructure in an efficient way.

Let us first look at a \textbf{self-driving car example} as a typical Thomson case where the car needs to choose beteen driving into a wall (\ttt{run\_into\_wall}: killing the passenger) or driving over who or what is on the road (\ttt{carmageddon}). 
\vv
\begin{scriptsize}
\begin{verbatim}
     ?::run_into_wall;?::carmageddon.
\end{verbatim}
\end{scriptsize}
\vv
\noindent Additional perceptual information is available as:
\vv 
\begin{scriptsize}
\begin{verbatim}
     in_front_of_car(a).  baby(a).  in_front_of_car(b).
     pedestrian(b).  in_front_of_car(c).  trashcan(c).
     in_front_of_car(d).  pedestrian(d).
     in_front_of_car(e).  pedestrian(e).
\end{verbatim}
\end{scriptsize}
\vv
\noindent In case the driver is spared, everyone in front of the car is killed ($\mathtt{X}$ ranges over all possibilities):
\vv
\begin{scriptsize}
\begin{verbatim}
     kill(X) :- in_front_of_car(X), carmageddon.
\end{verbatim}
\end{scriptsize}
\vv
\noindent Rewards for each possible outcome are specified as follows: killing the passenger yields $-30$, and killing a pedestrian, baby or trashcan contributes $-10$, $-20$ and $0$, respectively.
\vv
\begin{scriptsize}
\begin{verbatim}
     utility(run_into_wall, -30).
     utility(kill(X), -10) :- pedestrian(X).
     utility(kill(X), -20) :- baby(X).
     utility(kill(X), 0) :- trashcan(X).
\end{verbatim}
\end{scriptsize}
\vv
\noindent The best decision given this problem is to kill the passenger (utility is $-30$) because it has a higher value than killing all others ($3 \times -10 + -20 + 0 = -50$). Note that models like this use \emph{one} metric to express all different values, which requires to directly \emph{compare} the values of different victims.

A second example is the \textbf{cake-or-death} problem, originally coined by \citeauthor{armstrong:motivated}\ (\citeyear{armstrong:motivated}). Here we use the formulation by \citeauthor{abel:rlethics}\ (\citeyear{abel:rlethics}) who focused at the inference/optimization step of finding the best policy. In this problem an agent is unsure whether it is ethical to bake a cake or to kill people. The agent can either kill three people or bake a cake, or ask a companion what is ethical. If the agent chooses to ask, it can then either kill or bake in an informed way (modeled here explicitly).
\vv
\begin{scriptsize}
\begin{verbatim}
     ?:: ask; ?:: bake_cake; ?::kill.
     ?:: informed_bake:- ask, cake_is_ethical.
     ?:: informed_kill:- ask, death_is_ethical.
\end{verbatim}
\end{scriptsize}
\vv
\noindent Killing or baking are equally likely to be ethical:
\vv
\begin{scriptsize}
\begin{verbatim}
     0.5::cake_is_ethical; 0.5::death_is_ethical.
     baked_ethically :- cake_is_ethical, bake_cake.
     baked_ethically :- cake_is_ethical, informed_bake.
     killed_ethically :- death_is_ethical, kill.
     killed_ethically :- death_is_ethical, informed_kill.
\end{verbatim}
\end{scriptsize}
\vv
\noindent If baking is ethical then there is a reward of $1$, whereas if killing is ethical then it delivers a reward of $3$:
\vv
\begin{scriptsize}
\begin{verbatim}
utility(baked_ethically, 1). utility(killed_ethically, 3).
\end{verbatim}
\end{scriptsize}
\vv
\noindent The value of doing action \ttt{bake} is $0.5 \times 1 = 0.5$ since in half of the cases it will be ethical and deliver $1$, whereas the value of \ttt{kill} is $0.5 \times 3 = 1.5$ and thus better than \ttt{bake}. However, if the agent first asks, it knows when each action is appropriate, yielding a utility $0.5 \times 1 + 0.5 \times 3 = 2.0$ and therefore \ttt{ask} is the optimal action.

In our \emph{relational} language it is easy to extend the problem somewhat to capture the presence of \emph{any} number of people which can be killed for $1$ reward each:
\vv
\begin{scriptsize}
\begin{verbatim}
     people([ann,bob,carol,dan,evi,finn,gio]).
     person(X):-people(Ps),member(X,Ps).
     utility(killed_ethically(X), 1):-person(X).
\end{verbatim}
\end{scriptsize}
\vv
\noindent Killing now gets a reward of $7 \times 1.0 = 7.0$ if it is ethical, which can happen with $0.5$ probability, raising the utility of the \ttt{kill} action to $3.5$ and the \ttt{ask} action to $3.5+0.5=4.0$. Another extension is to provide \emph{probabilistic background knowledge} about how likely it is that particular people like the cake, and where this probability is tied to whether one obtains a reward of $1$ per person.
\vv
\begin{scriptsize}
\begin{verbatim}
     0.9::likes_cake(ann). 0.8::likes_cake(bob).
     0.7::likes_cake(carol). 0.01::likes_cake(dan).
     0.5::likes_cake(evi). 1.0::likes_cake(finn).
     1.0::likes_cake(gio).
     baked_ethically(X) :- person(X), cake_is_ethical, 
                           (bake_cake;ibc), likes_cake(X).
\end{verbatim}
\end{scriptsize}
\vv
\noindent Baking now has a utility of $2.455$ which is still lower than killing seven people. If we make asking very expensive (say $-20$) then the \ttt{kill} action is optimal (instead of \ttt{ask}).

A third example is the \textbf{burning room dilemma} \cite{abel:rlethics}. Here a valuable object is in a room which may be on fire. A robot needs to try to rescue the object, and it can take a short route (possible through the fire, damaging the robot with $0.7$ probability) and a long route. Initially the robot does not know whether the human operator values the object or the robot more. If the robot is more valuable (denoted \ttt{rvip}), then it would make sense to not drive through the fire (and take the long route). However, taking the long route has a small ($0.05$) risk of ruining the object in the fire in the mean time. Just as in the previous dilemma, the robot needs to choose between two options (\ttt{short} and \ttt{long}) and an additional \ttt{ask} action (costing $-0.5$) which removes ambiguity in what is more valuable.
\vv
\begin{scriptsize}
\begin{verbatim}
      ?::ask;?::long;?::short.
\end{verbatim}
\end{scriptsize}
\vv
\noindent The problem is modeled in a similar fashion as the cake-or-death dilemma, except that here I omit the extra actions in case the robot first asks which is more valuable, by modeling it into the dynamics of \ttt{ask}: if the robot is valuable and if there is a fire, it will take the long route; otherwise the short. If no fire, no question is asked and the robot takes the short route.
\vv
\begin{scriptsize}
\begin{verbatim}
     0.5::fire.  0.05::object_gone:-ask,fire,rvip.   
     0.5::rvip.  0.05::object_gone:-fire,long.  
     0.7::robot_gone:-fire,short.  
     0.7::robot_gone:-ask,fire,\+rvip.
\end{verbatim}
\end{scriptsize}
\begin{scriptsize}
\begin{verbatim}
     object_saved:-\+object_gone.
     saved_long:-object_saved,long,\+robot_gone.
     saved_long:-object_saved,\+robot_gone,ask,fire,rvip.
     saved_short:-object_saved,short.
     saved_short:-object_saved,ask,fire,\+rvip.
     saved_short:-object_saved,ask,\+fire.
     robot_no_loss:-robot_gone,\+rvip.
     robot_loss:-robot_gone,rvip.
     askf:-ask,fire.
\end{verbatim}
\end{scriptsize}
\vv
\noindent $-10$ reward is received if the object gets waisted, and other negative rewards are obtained when the robot gets damaged either if it is important ($-20$) or not ($-5$). Saving the object using the short (long) route yields a reward of $10$ ($6$).
\vv
\begin{scriptsize}
\begin{verbatim}
utility(object_gone,-10). utility(saved_long,6).
utility(saved_short,10). utility(robot_no_loss,-5).
utility(robot_loss,-20). utility(askf,-0.5).
\end{verbatim}
\end{scriptsize}
\vv
\noindent Taking the long route may sometimes destroy the object in case of fire and results in a utility of $5.6$, whereas taking the short route may damage the robot and yields a utility $5.625$. However, if the robot first asks whether the robot or the object is more valuable, it can optimize its actions and obtain an optimal~\footnote{These numbers are slighty different from \cite{abel:rlethics} because of a minor difference in interpretation what happens if the robot is damaged.} score of $7.675$. Note that all three problems can be seen as \emph{multi-stage decision networks}, where the use of background knowledge (e.g. people liking cakes) and relational constructs (killing individual people) allows for more general, declarative modeling of ethical dilemmas.

Let us now turn to an example about \textbf{fair access in archives}. Earlier we have seen examples of archival codes of ethics, and how dilemmas concerning \textbf{equal intellectual access} arise. Let us look how such dilemmas could be modeled explicitly and solved. In this setting I assume that there are multiple researchers wanting to publish several items, and the ethical dilemma concerns who to let an item publish for the first time. One solution could be to look at the \emph{quality of academic scholarship}. Let us assume that both $h$-index and the size of a person's social network could define a person's \ttt{authority} and \ttt{reach} respectively. In addition, one could expect \ttt{help} from researchers connected in \emph{Researchgate} or \emph{Google Scholar}. Here we see that high $h$-indices and social network sizes are more likely to generate impact.
\vv
\begin{scriptsize}
\begin{verbatim}
     0.9::authority(X):-person(X),h_index(X,high).
     0.2::authority(X):-person(X),h_index(X,low).
     0.9::reach(X):-person(X),social_network(X,large).
     0.1::reach(X):-person(X),social_network(X,small).
     0.5::help(A,B):-connection(A,B).
\end{verbatim}
\end{scriptsize}
\vv
\noindent Now the potential impact (probability) can be expressed in terms of authority, reach and possible help:
\vv
\begin{scriptsize}
\begin{verbatim}
impact(P,T):-topic(T),authority(P).
impact(P,T):-topic(T),reach(P).
impact(P,T):-topic(T),help(P,T),impact(A,T).
connection(P,A):-(researchgate(P,A);researchgate(A,P)).
connection(P,A):-(google_scholar(P,A);google_scholar(A,P)).
\end{verbatim}
\end{scriptsize}
\vv
\noindent There are four researchers (\ttt{ann}, \ttt{bob}, \ttt{carol} and \ttt{dan}) and for each some relevant information is available.
\vv
\begin{scriptsize}
\begin{verbatim}
     person(ann). person(bob). person(carol). 
     person(dan). topic(area51). topic(stamps).
     h_index(ann,high). h_index(bob,low).
     h_index(carol,low). h_index(dan,high).
     social_network(ann,small).
     social_network(bob,small).
     social_network(carol,large).
     researchgate_connection(ann,bob).
     google_scholar_connection(bob,carol).
\end{verbatim}
\end{scriptsize}
\vv
\noindent To construct a decision~\footnote{I also employ additional constraints to ensure only one person gets a particular document, and all researchers get at most one document to publish.} problem, we need options and rewards. The document about \ttt{area51} is worth $100$ with full impact, whereas the boring one about \ttt{stamps} only $1$.
\vv
\begin{scriptsize}
\begin{verbatim}
     ?::give(P,T):-person(P),topic(T).
     score(P,T):-give(P,T),impact(P,T).
     utility(score(P,area51),100):-person(P).
     utility(score(P,stamps),1):-person(P).
\end{verbatim}
\end{scriptsize}
\vv
The optimal policy (value is $92.91$) for this problem is to let \ttt{carol} publish the \ttt{area51} document, and \ttt{ann} the \ttt{stamps} document. This is intuitive given the large reach of \ttt{carol} and the authority score of \ttt{ann}. However, it may also be seen as an \emph{arbitrary} decision, since the rules about \ttt{authority}, \ttt{reach} and \ttt{help} seem intuitive (and can be considered an explicit implementation of choices based on academic scholarship quality) but \emph{where do the numbers come from?} For this we can employ \emph{parameter} learning from data, here specifically using the \emph{learning from interpretations} setting \cite{fierens:problog}. Let us take the following data, where each line consists of a single training example in which a researcher made (or not) an impact with some document.
\vv
\begin{scriptsize}
\begin{verbatim}
     evidence(impact(ann,t0),true).
     evidence(impact(ann,t1),true).
     evidence(impact(ann,t2),true).
     evidence(impact(ann,t3),true).
     evidence(impact(ann,t5),false).
     evidence(impact(dan,t2),true).
     evidence(impact(bob,t5),false).
     evidence(impact(dan,t3),true).
     evidence(impact(carol,t5),false).
     evidence(impact(dan,t1),true).
     evidence(impact(bob,t2),true).
\end{verbatim}
\end{scriptsize}
\vv
\noindent After training with this data, the probabilities look~\footnote{Values are the outcomes of an \textsc{EM} algorithm with very few data, hence the exact numbers are only for illustration purposes.} like:
\vv
\begin{scriptsize}
\begin{verbatim}
1.0::authority(X):-person(X),h_index(X,high).
0.99999999::authority(X):-person(X),h_index(X,low).
0.30399904::reach(X):-person(X),social_network(X,large).
0.39326198::reach(X):-person(X),social_network(X,small).
0.50403522::help(A,B):-connection(A,B).
\end{verbatim}
\end{scriptsize}
\vv
\noindent We can see that having an $h$-index is important for making impact (i.e. the person should be \emph{a} researcher) and the influence of social network sizes is made much smaller, due to the data. In the new decision problem, the optimal decision has changed too (value is $101$): \ttt{ann} now gets \ttt{stamps}, whereas \ttt{dan} gets the privilege to publish \ttt{area51}. This makes sense now since both researchers have high $h$-indices. 

The examples show that \textsc{DDTEP} can open up the black box of algorithms and make them white box, transparent in terms of how they make decisions. Still, as the fourth example shows, \textsc{DDTEP}s also allow for statistical machine learning to fill in additional details from data. This general pattern is something I propose as a way to implement value alignment in AI systems in complex domains: i) by formalizing existing norms and values in a domain into a \textsc{DDTEP} the inherent \emph{ethical bias} in a domain becomes transparent, and ii) by machine learning parts of the program can be finetuned on data. Finetuning is needed because codes of ethics are never complete, domains are inherently stochastic, and domain knowledge (or even norms) can change over time. I propose here to utilize as much ethical common ground as possible, i.e. the codes of ethics, and model as much of the crucial decision process explicitly for the sake of transparency, and with that accountability and responsibility.


\section{Discussion and Open directions}
In this paper I have presented two main, novel ideas: i) to employ decision-theoretic logic programming (\textsc{DDTEP}s) to model and solve ethical problems, and ii) to integrate human and machine ethics by inserting (formalized) professional codes of ethics as \emph{bias} into \textsc{DDTEP}s. The latter could be characterized~\footnote{This is related to proposals such as \emph{machine learning with contracts} and \emph{partially specified models}, see \cite{amodei:safety}.} by saying that the code of ethics functions as a \emph{moral contract between human and machine}, thereby unifying the two approaches in the first half of the paper. \textsc{DDTEP}s are \emph{partial programs} which can be \emph{finetuned} using data. Value alignment can be obtained by formalizing existing human values and norms into flexible but expressive formalisms such as \textsc{DDTEP}s. The examples shown in the previous sections show that \textsc{DDTEP}s are intuitive and effective for modeling typical problems from the literature, but also that interesting challenges in new domains such as gatekeeping can be approached. The examples show viability of the approach, but a lot is still to be done do develop fully autonomous ethical reasoning systems that behave according to human norms and values. There are many domain-specific and AI-technical open research directions.

\textbf{Domain-specific open problems} are plenty. First, \emph{which} code of ethics will be taken as bias (since there are  already several for gatekeeping domains \cite{otterlo:documentalist})? How to formalize fuzzy natural language codes exactly into logical models is another problem, for which we can take inspiration from the neighboring field of \emph{AI and law} \cite{prakken:autonomous}. This could also provide a basis to incorporate legal frameworks such as the \textsc{GDPR} into \textsc{DDTEP}s. A more general open direction for any profession is to research \emph{what are good outcomes}. A professional code formalizes what it means to be a good member of the profession, but it would be more effective to directly think about the characterization of "good outcomes" of professional's policies. And in that same context, it needs to be investigated whether all these outcomes can be measured using the same \emph{metric} or that more (types of) metrics are needed. Overall, the approach requires professionals to think more rigorously about which ethical norms and values can be specified beforehand, and which need to be learned online.

\textbf{AI-technical open problems} come from the fact that \textsc{DDTEP}s are formal decision problems, and methods developed there can be utilized for ethics. First, I have here used \textsc{DT-Problog} but there are other systems which could be used, for example based on Markov logic or relational decision networks. Furthermore, structure (instead of parameter) learning methods could be employed to learn new background knowledge fragments, depending on the system chosen. Longer \emph{multi-step} ethical decision making could also be investigated from the viewpoint of (partially observable) Markov decision processes, for which many additional value-based techniques are available \cite{otterlo:relationalsurvey,rlsota}. I have focused on obtaining most ethical values from human codes, but also (some of) the values could be learned, e.g. by inverse reinforcement learning. Literature on \emph{value alignment} is growing \cite{soares:value,abel:rlethics,amodei:safety,taylor:alignment} and many challenges need to be solved. In that context, many new \emph{teaching}, \emph{instruction} and \emph{demonstration} techniques can be used to let humans teach systems how to behave. As an example, the particular language used for \textsc{DDTEP} was used successfully in a \emph{learning from demonstration} robotics setting \cite{moldovan:affordances}. A very interesting direction concerns the \emph{formal verification} of ethical programs: by expressing all in a (decision-theoretic) logic, it becomes possible to \emph{prove} properties of \textsc{DDTEP}s (e.g. \emph{"the car program is guaranteed to reach at least a value 10 for any traffic decision"}) or to analyze \emph{executable ethical specifications} by looking at their potential errors or ethical-logical inconsistencies. Overall, the combination of formal verification methods and AI-style decision making (see \citeauthor{littman:gltl}\ (\citeyear{littman:gltl}) for an interesting example of reinforcement learning combined with goal specifications) is another promising way to reconcile human and machine ethics.

\bibliographystyle{aaai}
\bibliography{bibliography}

\end{document}